%% file: main.tex
\documentclass[10pt,twocolumn,letterpaper]{article}
\usepackage{cvpr}
\usepackage{times}
\usepackage{epsfig}
\usepackage{graphicx}
\usepackage{amsmath}
\usepackage{amssymb}
\usepackage{multirow}
\usepackage{array}
\usepackage{authblk}
\usepackage{romannum}
\usepackage{steinmetz}
\usepackage[pagebackref=true,breaklinks=true,letterpaper=true,colorlinks,bookmarks=false]{hyperref}

\newcommand{\Fig}{Fig. }
\newcommand{\Tab}{Table }
\newcommand{\Eq}{Eq. }

\newcommand{\prg}[1]{\par\noindent{{\textbf{#1}}}\\}

\cvprfinalcopy
\def\cvprPaperID{1710}

\ifcvprfinal\fi
\begin{document}
\makeatletter
\renewcommand\AB@affilsepx{, \protect\Affilfont}
\makeatother
\title{EventNet: Asynchronous Recursive Event Processing}
\author[$\dagger$]{Yusuke Sekikawa}
\author[$\dagger$]{Kosuke Hara}
\author[$\ddagger$]{Hideo Saito}
\affil[$\dagger$]{Denso IT Laboratory}
\affil[$\ddagger$]{Keio University}

\maketitle
\begin{abstract}
Event cameras are bio-inspired vision sensors that mimic retinas to asynchronously report per-pixel intensity changes rather than outputting an actual intensity image at regular intervals.
This new paradigm of image sensor offers significant potential advantages; namely, sparse and non-redundant data representation.
Unfortunately, however, most of the existing artificial neural network architectures, such as a CNN, require dense synchronous input data, and therefore, cannot make use of the sparseness of the data. 
We propose EventNet, a neural network designed for real-time processing of asynchronous event streams in a recursive and event-wise manner.
EventNet models dependence of the output on tens of thousands of causal events recursively using a novel temporal coding scheme.
As a result, at inference time, our network operates in an event-wise manner that is realized with very few sum-of-the-product operations---look-up table and temporal feature aggregation---which enables processing of 1 mega or more events per second on standard CPU.  
In experiments using real data, we demonstrated the real-time performance and robustness of our framework. 
\end{abstract}

\input{section/intro}
\input{section/EventNet}
\input{section/experiment}
\input{section/related}
\input{section/conclution}
\newpage
\clearpage
{\small
\bibliographystyle{ieee}
\bibliography{bib}
}
\end{document}

%% file: section/intro.tex
\input{fig/_overview}
\section{Introduction}
Existing frame-based paradigms---dense synchronous video stream acquisition and dense/batch processing---cannot scale to higher frame rates or finer temporal resolutions because the computational complexity grows linearly with the processing rate or temporal resolution (\Fig \ref{fig:overview} top).
The grows of the computational complexity comes from the redundant synchronous measurement/transmission of dense intensity frames for unchanged pixels and the following redundant signal processing algorithm, such as convolutional neural networks (CNNs) \cite{tran2015learning, krizhevsky2012imagenet,girshick2014rich,ren2015faster,zhang2016faster}, which computes the sum of the products,  even for the unchanged pixels. 
Furthermore, the same frames are computed multiple times (temporal sliding window operation) to model temporal dependencies \cite{tran2015learning}.

The event-based camera \cite{4444573} discards the frame-based paradigm and instead adopts a bio-inspired approach of independent and asynchronous pixel brightness change measurement without redundancy. 
This new type of data acquisition has the potential to enable a new paradigm of high-speed, non-redundant signal processing using the naturally compressed non-redundant  event stream. 

Our research goal was to develop a neural network architecture that can process an extremely high-rate,\footnote{Maximum event rate of the iniVation DAVIS240 camera is $12$ mega events per second.} variable length, and non-uniform raw event stream in real time.
To this end, we proposed EventNet, a neural network designed for the real-time processing of an asynchronous event stream in an event-wise manner.
Our main contributions are summarized as follows:

\prg{Recursive Architecture}
We proposed a recursive algorithm by formulating dependence on causal events  (which could be tens of thousands)  to the output recursively using a novel temporal coding and aggregation scheme that comprises a complex phase rotation and complex $\max$ operation. 

\prg{Lookup Table Realization of MLP}
The deep multi-layer-perceptron (MLP) appears in the recursive formula and dominates most of the computation.
It was replaced by a lookup table (LUT) at inference time by the factorization of the temporal term. 
This replacement removed most of the sum-of-product operations of the MLP.

\prg{Asynchronously Two Module Architecture}
The entire network was separated into two modules working asynchronously; 
an event-driven module that updates the global feature immediately as it receives a new event, and the on-demand module that computes the final output with a lightweight MLP. 
This separate architecture avoids a wasteful computation of output that is not used by applications (\Fig \ref{fig:overview} bottom). 

\prg{Applicability to Real-World Data}
We demonstrated the applicability of EventNet for real-world applications using publicly available datasets.
We applied EventNet to event-wise semantic segmentation, object motion estimation, and ego-motion estimation. 
These demonstrated real-time performance in the CPU. 
The event-driven module---comprising an event-wise LUT, a temporal code, and $\max$---worked extremely fast and it could process about $1$ mega event per second (MEPS)  on a standard CPU. 
Further, the on-demand inference module was capable of responding to a request from an application at 1 kHz or more on the CPU.

%% file: fig/_overview.tex
\begin{figure*}[!h]
\begin{center}
\includegraphics[width=2\columnwidth]{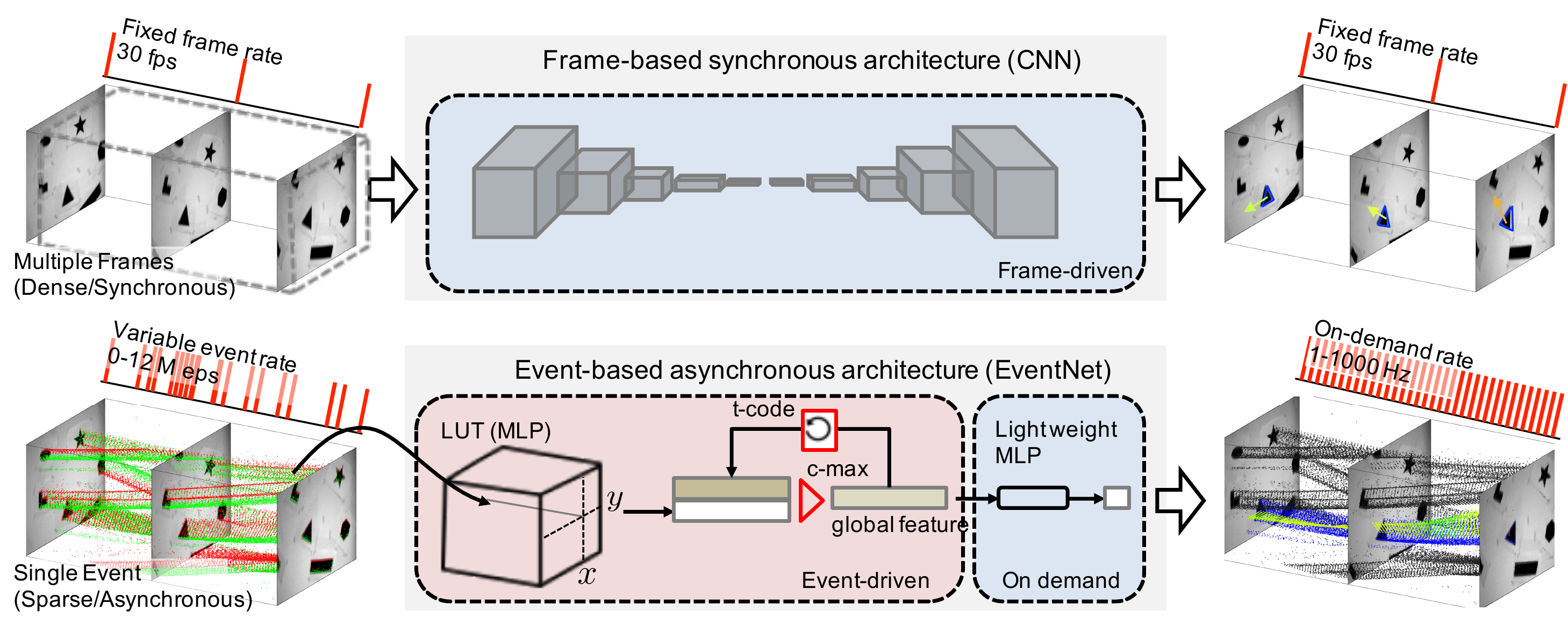}
\end{center}
\caption{
\label{fig:overview}
\textbf{Overview of asynchronous event-based pipeline of EventNet (inference) in contrast to the conventional frame-based pipeline of CNN.}
Top: Conventional frame-based paradigm (CNN). 
Bottom: Proposed event-based paradigm  (EventNet).
In the conventional frame-based paradigm, the computation is redundant, which prevents the algorithm from scaling to higher frame rates.
Due to the slow output rate, applications are obliged to use old results. 
Our EventNet can directly process sparse events in an event-wise manner without densification; it processes only changed pixels, and thus, there is no redundancy.
When a new event arrives, global features (summary of current states)  are updated with a highly efficient recursive operation using an LUT at the event rate.
And, when the application requests output, it is computed on demand by feeding the global-features to a lightweight MLP.
This pipeline is extremely efficient and can process more than  $1$ mega events per second (MEPS); it can also respond to the on-demand request at 1 kHz or more with a standard CPU.
Examples of input event streams are shown (\textit{green} and \textit{red} indicate positive and negative events, respectively) on the right, and the results of the global object motion estimation and event-wise semantic segmentation are shown on the right (events classified as triangles are shown in \textit{blue}, and others are shown in \textit{grey}, and the estimated motion vector of a triangle is shown as an arrow where the color encodes the angle of the motion). The data from \cite{EventData} are used for this experiment.
Notes: MLP = multi-layer perceptron, t-code = temporal coding, and c-max = complex max pooling.
}
\end{figure*}

%% file: section/EventNet.tex
\section{EventNet}
\label{sec:hougtnet}
\input{fig/_network}

\subsection{Event-Based Camera}
Each pixel of an event-based camera asynchronously measures intensity levels and reports an event---$(x,y,p,t) $ pixel location, polarity indicating the positive or negative changes in intensity, and time stamp---when changes in intensity are detected. 
We considered a sequence of events within a $\tau$ ms interval based on the time stamp $t_j$ of the $j$-th event as $\mathbf{e}_{j}:=\{e_i|i=j-n(j)+1,...,j\}$, where each event $e_i$ is a quartet of its $(x_i, y_i, p_i, \Delta t_{j,i})$, where $\Delta t_{j,i}$  represents the time difference $\Delta t_{j,i}=(t_j-t_i)$, 
and $\mathbf{e}_{j}$ is updated when a new $(j+1)$-th event arrives by adding $e_{j+1}$ and removing events that come out of the $\tau$ ms interval.
Thus the length of the sequence $n(j)$ changes dynamically (\Fig \ref{fig:batch} right). 

\subsection{Problem Statement}
\label{seq:problemstatement}
We considered the function $f$, realized by neural networks,  to estimate target value  $y_{j}$  given event stream $\mathbf{e}_{j}$: $y_j=f(\mathbf{e}_{j})$.
Events come at a nonuniform rate, ranging from $0$ to millions per second,  and the network needs to process the variable-rate data in an event-wise manner.
Our goal was to realize a trainable event-driven neural network  $f$ that satisfies all of the following conditions: 

\prg{\romannum{1}) End-to-End Trainable}
To realize the practical supervised learning of applications, the network needs to be  trainable end-to-end by using error backpropagation (BP) using a supervised signal.

\prg{\romannum{2}) Event-Wise Processing}
The events are spatially sparse and temporally nonuniform. 
The dependence of output $y_{j}$ on the sparse signal  $\mathbf{e}_{j}$ needs to be processed in an event-wise manner  without densification to voxel representation to avoid redundant computation.

\prg{\romannum{3}) Efficient Recursive Processing}
The event rate could be very high (more than 1 MEPS), depending on the scene, and thus, the size of  $\mathbf{e}_{j}$ could be large (tens of thousands depending on the event rate and window size $\tau$). 
It is quite impossible to process such a large variable length nonuniform event-stream $ \mathbf {e} _ {j} $ in a batch manner at a very high event rate.
Thus, an efficient algorithm that can recursively process the stream is required.

\prg{\romannum{4}) Local Permutation Invariance}
An ideal event-stream recognizer  $f$  needs to be invariable against data-point permutations in a \textit{short time} window, while having sensitivity to capture \textit{longer time} behaviors.
Invariance against a temporally local permutation is essential because the order that event data emerges in a temporal vicinity can be generally thought of as stochastic due to the limited temporal resolution (e.g., $1\,\mu$s) and the noise of time stamps (\Fig \ref{fig:batch} left).
Thus, the order of incoming events may change even if the same scene is observed with the same camera motion.
On the other hand, the  long-range temporal evolution of an event needs to be modeled to capture motion in a scene.

\subsection{Symmetric Function}
To cope with the permutation, a simple MLP or RNN with randomly permuted sequences is not a feasible choice because it is difficult to scale to thousands or tens of thousands of input elements \cite{qi2017pointnet}. In addition, it is impossible to be totally invariant to the permutation \cite{2015arXiv151106391V}. 
The PointNet architecture \cite{qi2017pointnet} solves the problem in a theoretical and concise way by approximating the function  as
 \begin{equation}
 \label{eq:pointnet}
y_{j}=f(\mathbf{e}_{j})\approx g(\max(h(e_{j-n(j)+1}),...,h(e_{j}))),
 \end{equation}
where , $h:  \mathbb{R}^4 \rightarrow \mathbb{R}^K$, $\max:  \underbrace{\mathbb{R}^K\times ...\times  \mathbb{R}^K}_{n(j)} \rightarrow \mathbb{R}^K$,  and $g:  \mathbb{R}^K \rightarrow \mathbb{R}$. 
They approximate $h$ and $g$ using an MLP.
Because of the symmetric function $\max$, the permutation of events does not change the output $y_{j}$.
Note that $\max$ operates independently for each dimension of $\mathbb{R}^K$.

\subsection{EventNet}
PointNet focuses on processing sets of vectors such as a 3D point cloud in a batch manner. 
When we attempt to use  PointNet to sequentially process a stream of events, a huge amount of computation of  $h$ realized by the MLP and $\max$ makes real-time processing difficult:
For all $n(j)$ events in $\mathbf{e}_{j}$, when a new $(j+1)$-th arrives,   the time difference  $\Delta t_{j,i}=(t_j-t_i)$ changes to $(t_{j+1}-t_i)$; thus, most of the $n(j)$ event that has already been processed  by $h$ (realized by deep MLP) needs to be processed again by $h$ when we receive the new event as long as it is within ther $\tau$ ms time window.
If the function $\max$ is the function of the set of  $n(j)$ high-dimensional feature vectors and the vector changes with the same situation as above, we need to compute $\max$ within all  $n(j)$  feature vectors at the event rate.
The single cycle of these computations are themselves intensive since $n(j)$ may be thousands or tens of thousands in common scenarios.
Furthermore, these two computations should run on the event rate (could be more than $1$ MEPS).
These issues make it impossible to use PointNet to process event streams in real time.

To overcome the above difficulty in processing the event streams, we proposed EventNet, which processes the sparse signal recursively rather than processing large numbers, $n(j)$, of events in batch manner.
 
\prg{Temporal Coding}
Since the function $h$ is a function of time difference $\Delta t$, the network is required to compute $h$ tens of thousands of times for the same event (as long as the event exists within the time window) as  the new event is being received.
Simply removing $\Delta t$ from the input $e$ creates a loss of important temporal information, resulting in a deterioration of the performance (this will be discussed in Section \ref{sec:ablation}).
To avoid the multiple time computations of $h$ for the same events while keeping the temporal information, we removed the dependence on $\Delta t$ from $h$ and instead introduced a temporal coding function $c$ to encode the information of $\Delta t$ as $h(e_i)=c(h(e^-_i),\Delta t_{j,i})$, where  $e^{-}:=(x,y,p)$. Then \Eq \ref{eq:pointnet} becomes
\begin{equation}
 \label{eq:eventnet1}
f(\mathbf{e}_{j})\approx g( \max(c(z_{j-n(j)+1},\Delta t_{j,j-n(j)+1}),...,c(z_{j},0))),
 \end{equation}
where, $z_i=h(e^-_i) \in \mathbb{C}^K$. 
Using this formulation, we need to compute $h$  only once for each observed event; however, $c$ and $\max$ need to be computed for all events in the time window every time a new event arrives.  

\prg{Recursive Processing}
\label{sec:recursive}
We considered processing $\mathbf{e}_{j-1}$ and $\mathbf{e}_{j}$ sequentially.
Let the time difference of  latest time stamp be $\delta t_j:=t_{j}-t_{j-1}$, and assume the norm of each element of $z_i$ is less than $1$ (by $\tanh$). 
We want to make \Eq \ref{eq:eventnet1} recursive, cf, computing $\max$ at $j$ using $\max$ at $j-1$ and event $e_j$.
For this, the composition of $\max$ and $c$ need to be recursive. 
The $\max$ is not recursive for general time series vectors  unless the window size is $\infty$,
so we propose temporal coding function $c$ of \Eq \ref{eq:tcode}, which guarantees recursiveness:
\begin{equation}
\label{eq:tcode}
a_{j,i}=c(z_i,\Delta t_{j,i})=\left[|z_i|-\frac{\Delta t_{j,i}}{\tau}\right]^{+}\exp\left(-\mathrm{i}\frac{2\pi \Delta t_{j,i}}{\tau}\right),
\end{equation}
where the first term decays the input linearly to the elapsed time, and the second term encodes the temporal information by complex rotation.
As $c$ decays the input linearly on elapsed time (constant decay within the fixed time window), it satisfies the relation $\Delta t_{j,i}=\sum_{k=i,...,j}\delta t_k$.
The norm of each element of $a$ that is older than $\tau$ is always zero, and the complex rotation of each feature represents elapsed time  by sequentially rotating the element by $2\pi\delta t_j/\tau$. 
This coding function makes the composition of  $\max$ and $c$ recursive for a finite window size,
\begin{multline}
\label{eq:reqmax}
\max((c(z_{j-n(j)}, \Delta t_{j-n(j)}),..., c(z_{j+1}, 0)))\\
=\max(c(s_{j},\delta t_j), z_{j+1}),
\end{multline}
where $s_{j}$ is the result of $\max$ at time $t_j$. 
Putting \Eq \ref{eq:tcode} into \Eq \ref{eq:eventnet1}, we get the following recursive algorithm:
\begin{equation}
\label{eq:recursive_2}
f(\mathbf{e}_{j+1}) \approx  g(\max(c(s_{j},\delta t_{j}),h(e^-_{j+1})),
\end{equation}
where global feature $s_{j}:=\max(c(s_{(j-1)},\delta t_{j-1}),h(e^-_{j}))$ is updated  recursively.
This formulation has favorable characteristics for sequential event processing: \romannum{1}) we only need to compute $h(e^-_i)$ once for each event; \romannum{2})  $c$ is computed only for $s$, not for all $n(j)$ vectors; and \romannum{3})  $\max$ is computed between only  two vectors instead of between all $n(j)$ vectors. 
A permutation of events that has the same time stamp does not change the result as desired, and because the coding function of \Eq \ref{eq:tcode} is approximately constant within a small temporal interval, small perturbations of time stamps from noise will result in small changes in output $y_j$ as long as the function $g$ is smooth.

Note that $\max$  in the above equation is defined in a set of complex values. Let  $a^k_1,...,a^k_n\,|\,a^k_i\in \mathbb{C}$ be a  $k$-th channel of temporal sequence (i.e., sequence of feature vectors after the temporal coding), then the complex $\max$ is defined as;
\begin{equation}
\label{eq:complexmax}
\max(a^k_1,...,a^k_n)=a_i\textit{, where}\,\, i=\arg\max_i (|a^k_i|).
\end{equation}

\prg{LUT Implementation of MLP}
Because the spatial position and polarity from event-based cameras are discrete, there are only $W\times H\times 2$ patterns in inputs $e^-$ (spatial position and polarity).
Therefore, we can precompute  results of $h$, and utilize LUT in inference time for computing high-dimensional vectors. 
This is considerably faster than Deep MLP $ h $, which contains a large number of product-sum operations.

\prg{Summary}
EventNet satisfies all the conditions described in Section \ref{seq:problemstatement}.
It is trainable using BP in supervised manner because it uses differentiable MLP, it can efficiently process sparse event signals without densification  with a novel recursive architecture of \Eq\ref{eq:recursive_2}.
It can also capture a long-range temporal evolution by a complex rotation while also being invariant to the small  temporal vicinity due to the max operation.

\subsection{Network Architecture}
The EventNet model of \Eq\ref{eq:recursive_2} is realized in the architecture shown in \Fig \ref{fig:network}.
The function $h$ is realized as mlp1 and mlp2, the function $g$ is realized as mlp3 for global-estimations, and mlp4 for event-wise estimation.

Depending on the application, the required rate of output varies, and computing a final output at the event rate ($1$ MEPS) is a waste of computation as most of the results are not used by the application.
In addition, 1000 Hz may be more than enough for many applications.
To achieve real-time event processing and high-rate estimation without wasteful computation, our network comprises two separate modules that work asynchronously with each other.

\prg{Event-Drive Processing Module}
This module operates in an event-driven manner: when a new event $e_j$ arrives at the network asynchronously, it is immediately processed to update the global feature vector $s_j$, 
which is realized by the recursive algorithm of \Eq \ref{eq:recursive_2}.
Furthermore, mlp1 and mlp2 are realized as an LUT, which is much faster than feed-forwarding deep MLP.

\prg{On-Demand Processing Module}
This module operates on demand from an application. When the application requires the latest estimation, the output is computed on demand with mlp3 or mlp4.
Because the input to mlp3 is a single vector, its computation is reasonably fast, and 1000 Hz with a standard CPU is easily achieved. 

\prg{Asymmetric Training/Inference}
As shown in \Fig \ref{fig:network} the network structure of EventNet differs in training and inference; it utilizes a batch structure when training.
In principle, EventNet can be trained recursively using \Eq\ref{eq:recursive_2} as long as batch normalization (BN) is absent. 
The recurrent structure does not allow for the computation of batch statistics. 
However, we adopt the batch structure during training for two reasons: \romannum{1}) to use BN, which plays a very important role, based on our experience; and \romannum{2}) to parallelize the computation of MLP for  efficiency.
The structure of EventNet when training are similar to PointNet \cite{qi2017pointnet} except for the temporal coding layer (\Fig \ref{fig:network} middle), and they are trained in largely the same way.\footnote{In terms of implementation, they are greatly different because EventNet needs to handle variable-length data The implementation details are described in the supplemental material.}
However, the temporal coding makes a huge structural difference and a large computational gain in inference time (\Fig \ref{fig:network} bottom, \Tab \ref{table:speed}).

%% file: fig/_network.tex
\begin{figure*}[!h]
\begin{center}
\includegraphics[width=2\columnwidth]{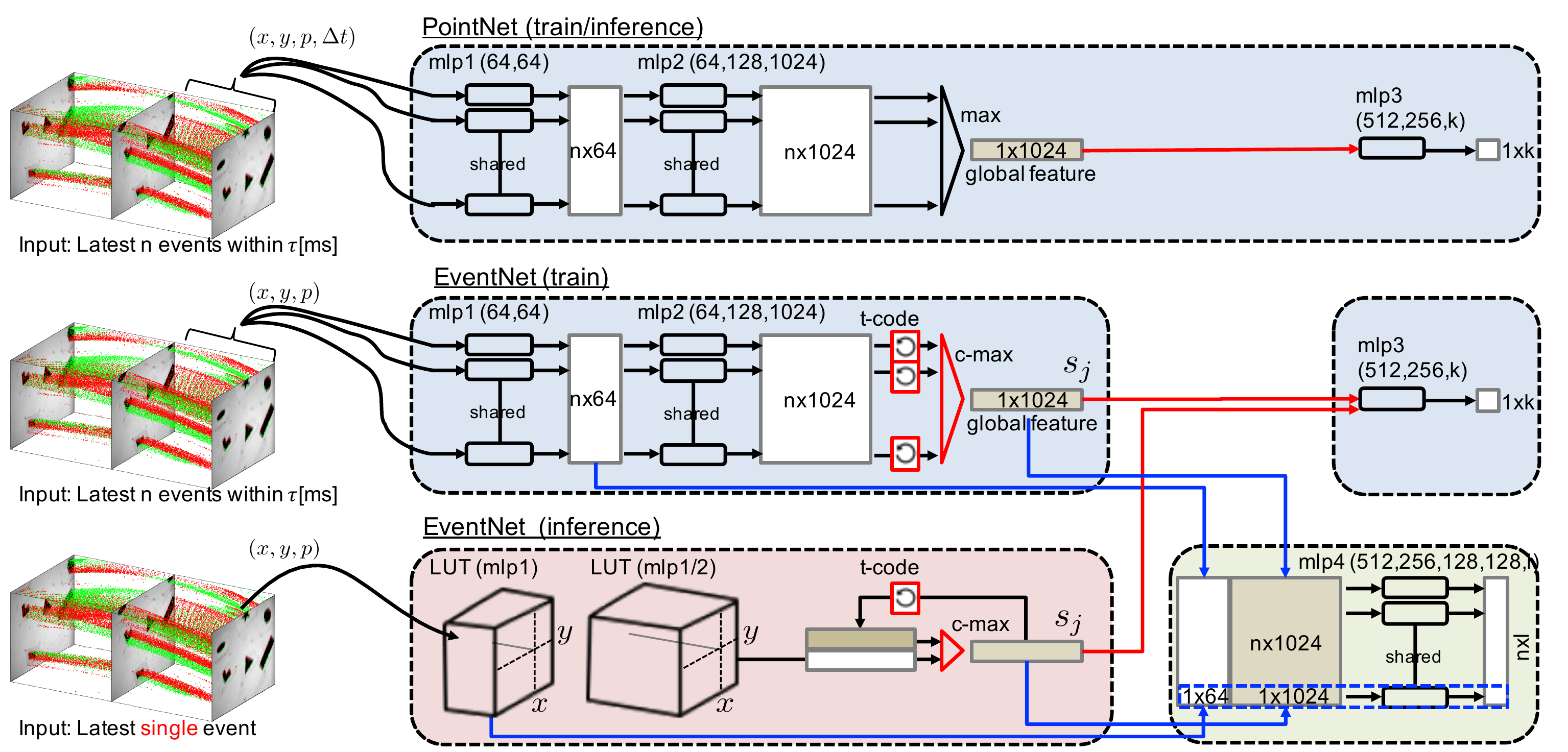}
\end{center}
\caption{
\label{fig:network}
\textbf{EventNet architecture}
The network architecture of our EventNet is shown in comparison with PointNet.
Our network has the novel temporal coding layer of \Eq \ref{eq:tcode}.
Thanks to this layer, the dependence on the sequence of events is computed recursively.
Furthermore, the most computationally significant part (mlp1 and mlp2, which are trained using standard error backpropagation) are realized as a lookup table (LUT) after training, which is significantly faster than an MLP.
As a result, EventNet processes streams of events efficiently in an event-driven manner---compute per-event feature by LUT,  apply the temporal code to the global feature, and aggregate global feature by max pooling of the two vectors---which is repeated recursively as it receives a new event.
Numbers in bracket are layer sizes. 
Batch normalization \cite{Ioffe:2015:BNA:3045118.3045167} is used for all layers except output layers. 
Similar to PointNet, EventNet has a variant of architecture that can output on a per-event basis, which is realized with mlp4 by a concatenation of local feature and global feature(\textit{blue} line).
}
\end{figure*}

%% file: section/experiment.tex
\section{Experiments}
\label{sec:experiments}
\input{table/accuracy}

\input{table/speed}

The purpose of the experiments was to evaluate the computational efficiency and robustness of EventNet in a practical application. 
To the best of our knowledge, there are no end-to-end trainable neural networks capable of processing raw event streams in real time.
Nevertheless, we compare EventNet with PointNet as it is a successful method that can directly process reasonably good-sized point set. 
We want to emphasize that because PointNet cannot process event streams in a recursive manner, as EventNet does, it is thousands of times slower (depending on the event rate) than EventNet at such processing and cannot process an event stream in real-time.

We will first describe the learning procedure (Section \ref{sec:learning_procedure}) and datasets used for the experiments (Section \ref{sec:dataset}).
Next, we demonstrate the real-time processing capability of  EventNet in several real-world tasks (Section \ref{sec:applications} and Section \ref{sec:quantitative}).
And last, we provide the results of an ablation study to reveal the effects of core components of EventNet (Section. \ref{sec:ablation}).  

\subsection{Learning Procedure}
\label{sec:learning_procedure}
\prg{Mini-Batch Construction}
\input{fig/_batch}
Training MLPs in EventNet requires processing events streams in a batch manner (\Fig \ref{fig:network} EventNet (train)), although it can process event streams recursively in inference time. 
Given an entire event-stream for training, the single event stream $\mathbf{e}_j$ for training was composed as follows.
An event that corresponds to the latest event within the temporal window is randomly selected from the entire event stream (let the index of the event be $j$).
Then, cut out an event stream on a $\tau$ ms interval based on $t_j$, $\mathbf{e}_{j}:=\{e_i|i=j-n(j)+1,...,j\}$ (\Fig \ref{fig:batch}).
Optionally, to increase the variety in training data, it can then be randomly spatially cropped.
Note that the length of each sequence $n(j)$ is different since event rates change depending on the scene or the motion of the camera.

\prg{Optimization}
All networks, including the one from the ablation study, were trained using the same protocol.
Single-epoch consisted of randomly composed  $8,000$ event stream, and the training was carried out for 500 epochs.
For the optimization, we used Adam \cite{kingma2014adam} with the recommended hyper-parameter settings of: $\beta_{1}=0.9$, $\beta_{2}=0.999,$ and $\epsilon=10^{-8}$.
The initial learning was $0.0002$, and it was divided by 2 every 20 epochs until epoch 100, after which it was kept constant. 
The decay rate for BN started at 0.5 and gradually increased to 0.99. 
All implementations used the MatConvNet \cite{vedaldi15matconvnet} library, and we carried out all our training on a single NVidia V100 GPU. 

\subsection{Datasets}
\label{sec:dataset}
We used publicly available datasets captured with real event-based cameras for evaluation.
\prg{ETHTED+}
The first and second applications used the datasets from \cite{EventData}  with our additional  hand-annotated segmentation label\footnote{The additional label will be published with this paper.} for \textit{shape rotation sequence} (ETHTED+).
The supplemental annotation is given on a per-event basis, indicating whether it came from the triangle or not.
The camera used for this dataset was the DAVIS 240, which has an array size of $180\times 240$.
Each training event stream was randomly cropped spatially to $128\times 128$ from the original  $180\times 240$ to increase the variety of spatial displacements. 
The average event rate of this data set (center $128\times 128$ region)  was $0.16$ MEPS.
We used the first $50$ seconds of the sequence for training, and the other 10 seconds was used for testing.
The temporal window size of $\tau=32$ ms was used for this dataset.

\prg{MVSEC}
The third application used  MVSEC \cite{MSECD}, consisting of event sequences captured in road scenes, which are much larger than ETHTED+; thus, it had more variation in the input data space. 
The camera used for this dataset was the DAVIS 346, which has an array size of  $260\times 346$.
The average event rate of this data set was $0.35$ MEPS.
To remove noise, we applied the nearest neighbor filter followed by refractory period filter, as described in  \cite{10.3389/fnins.2018.00118}, where we used $5$ ms and $1$ ms as temporal window size respectively.
We used \textit{outdoor day1}, \textit{outdoor day2}, \textit{outdoor night1}, and \textit{outdoor night2} sequences for training and used \textit{outdoor night3} for testing.
A temporal window size of $\tau=128$ ms was used for this dataset.

\subsection{Applications}
\label{sec:applications}
We demonstrated the applicability of  EventNet for several real-world tasks.
\input{fig/_quolitative}

\prg{Target Motion Estimation}
In this application, the network estimates the motion of a specific object (triangle) using global-estimation network (see \textit{red} line in \Fig\ref{fig:network} with mlp3).
We used  ETHTED+ for this experiment.
Using the class label of events, we computed the motion $[u,v] $ of the triangle by linearly fitting the centroid position of events within $33$ ms intervals.
The input event stream was processed to compute global features at the event rate, and the target values $[u,v] $ were computed on demand at 1000 Hz using the global feature. 
When testing, the cropping region was fixed to the center region.

\prg{Semantic Segmentation}
In this application, the network estimates the class label of each event (triangle or others) using event-wise-estimation network (see \textit{blue} line in \Fig\ref{fig:network} with mlp4).
We used ETHTED+ for this experiment.
The input event stream was processed to compute per-event features from mlp1 and global features at the event rate.
The target values which was the probability of each class, was computed on demand using concatenation of global features and local features. 
We note that temporal information may be less important for this task  because we can still determine the shape from the event stream even if the temporal information was lost.
The network for this application was trained jointly with the global-estimation network sharing mlp1/2.
An example of processing results for object-motion estimation and semantic segmentation is shown in \Fig \ref{fig:overview}.

\prg{Ego-Motion Estimation}
In this application, the network estimates ego-motion of the camera using global-estimation network.
We used  MVSEC for this experiment and the yaw-rate, which was provided with the dataset was used as a target values.
The qualitative results are shown  in \Fig \ref{fig:experiments_example}.

\subsection{Quantitative Comparison}
\label{sec:quantitative}
Estimation accuracy for the three applications are reported in \Tab \ref{table:accuracy} on top, and computational times are reported in \Tab \ref{table:speed}.
In summary,  EventNet achieved a performance that is comparable to PointNet while realizing less than 1 $\mu$s processing for a single event,  proving it can achieve real-time processing up to 1 MEPS, which covers most practical scenarios.
\prg{Estimation Accuracy}
As shown in  \Tab \ref{table:accuracy}, our model achieved comparable performance to PointNet in all three of the experiments, while also achieving real-time performance using much less memory usage.
For the object motion estimation task, we conducted preliminary experiments to see the effects of different time windows of 8, 16, and 32 ms.
The largest (32 ms) performed the best with our EventNet, but the performance of PointNet was almost the same across the different time windows.
The results and a discussion of the  results  are reported in the supplemental material. 

\prg{Computation Time}
We assumed the system displayed in \Fig \ref{fig:overview} for the comparison shown in \Tab \ref{table:speed}.
The required number of operations for the MLP and $\max$ was about $n(j)\times$ less than processing in batch manner due the novel recursive formula used in EventNet. 
Furthermore, the LUT realization of mlp1/2  improved the computation speed by about $45\times$.
Consequently, EventNet processed input event rates of $1$ MEPS with standard CPU, covering most of the practical scenarios.
Conversely, PointNet cannot process event streams recursively and is, thus, required to process $n(j)$ events within a $\tau$ ms time window in a batch manner every time it receives a new event by reprocessing the events again and again. 
Real-time processing is, therefore, entirely impossible.
Note that the per-event computation time of EventNet is not affected by the event rate, while the computational complexity grows linearly with the rate in the case of PointNet.

Further, EventNet is much more memory efficient than PointNet  because it requires storage of only one global feature vector as it processes incoming event streams recursively (\Eq \ref{eq:recursive_2}); and it operates with an LUT and is, thus, not required to store intermediate feature maps.

\subsection{Ablation Study}
\label{sec:ablation}
In this section, we discuss our study of the contribution of temporal coding of \Eq \ref{eq:tcode}, the key component of EventNet, which enables highly efficient recursive event-wise processing.
For this, we ran the object motion estimation and semantic segmentation experiments using ETHTED+, which is the same as the ones discussed in Section \ref{sec:applications}.
We examined the contribution of the temporal decay term $([|z_i|-\frac{\Delta t_i}{\tau}]_{+})$ (TD) and the complex temporal rotation term $\exp(-\mathrm{i}\frac{2\pi \Delta t_i}{\tau})$ (TR) from the equation.
The results are summarized at the  bottom in \Tab \ref{table:accuracy}.
Because segmentation accuracy was equally good for all variants, we discuss the performance in terms of the  object motion estimation accuracy,
in which the temporal information may be more informative for the estimation.

\prg{EventNet Without Temporal Decay}
This variant disabled only decay term.
Without the decay, the recursion of \Eq \ref{eq:recursive_2} was not satisfied; thus, it is necessary to compute $\max$ for all $n(j)$ vectors.
Therefore, this variant cannot process $1$ MEPS of events in real-time.
The estimation accuracy was the best among all variants, including full EventNet.

\prg{EventNet Without Temporal Rotation}
This variant disabled only the complex temporal rotation term.
Because this variant has linear decay term, the recursion of \Eq \ref{eq:recursive_2} was satisfied; thus, it can process event streams in an event-wise manner. 
Actually, this variant is slightly faster than the full EventNet because it does not compute complex rotation.
The performance was not as good as EventNet or EventNet without TD, which may be attributed the lack of temporal information without explicit coding of temporal information as complex rotation.

\prg{EventNet Without All}
This variant disabled both terms.
As a result, the network structure of this variant was the same as PointNet, and the difference to PointNet was that this variant did not include temporal term $\Delta t$ as input.
This variant cannot operate in real time for the same reason that EventNet without TD cannot.
This variant showed the worst performance of the variants, which may explained by this variant not having any temporal information.

%% file: table/accuracy.tex
\begin{table*}
\caption{\label{table:accuracy}$\textbf{Quantitative evaluation using ETHTED+ and MVSEC.}$
Quantitative evaluations of our EventNet and  PointNet \cite{qi2017pointnet} are shown.
For experiments using ETHTED+, we report global accuracy (GA) and mean intersection of union (mIoU) for semantic segmentation and L2 regression error for object motion.
For experiments using MVSEC, we evaluated the L2 regression error for  ego-motion estimation.
Our EventNet achieved a comparable performance to PointNet while achieving real-time performance.
On the bottom, the results of EventNet when disabling each term of temporal coding in \Eq \ref{eq:tcode} are shown (see main text for the abbreviation).
The number in parentheses indicates the standard deviation of the results.
}
\begin{center}
\begin{tabular}{cc|c|c|c|c|c}
\hline 
\multicolumn{2}{c|}{} & \multicolumn{3}{c|}{ETHTED+} & MVSEC & \multirow{3}{*}{%
\begin{tabular}{c}
Real-time\tabularnewline
processing at $1$ MEPS\tabularnewline
\end{tabular}}\tabularnewline
\cline{3-6} 
\multicolumn{2}{c|}{} & \multicolumn{2}{c|}{Semantic segmentation} & Object-motion  & Ego-motion & \tabularnewline
\cline{3-6} 
\multicolumn{2}{c|}{} & GA {[}\%{]} & mIoU {[}\%{]} & error {[}pix/$\tau${]} & error {[}deg/sec{]} & \tabularnewline
\hline 
\multicolumn{2}{c|}{PointNet} & $98.9$ & $97.4(0.13)$ & $3.14(0.08)$ & $4.55$ & NO\tabularnewline
\hline 
\multicolumn{2}{c|}{EventNet} & $99.2$ & $97.5(0.22)$ & $3.11(0.28)$ & $\textbf{4.29}$ & YES\tabularnewline
\hline 
\hline 
\multirow{3}{*}{Ablation} & w/o TD & $\textbf{99.4}$ & $\textbf{98.8}(0.16)$ & $\textbf{3.08}(0.32)$ & \multirow{3}{*}{\textemdash} & NO\tabularnewline
\cline{2-5} \cline{7-7} 
 & w/o TR & $98.1$ & $97.9(0.11)$ & $3.74(0.06)$ &  & YES\tabularnewline
\cline{2-5} \cline{7-7} 
 & w/o ALL & $98.3$ & $97.1(0.25)$ & $4.14(0.32)$ &  & NO\tabularnewline
\hline 
\end{tabular}
\end{center}
\end{table*}

%% file: table/speed.tex
\begin{table*}
\caption{\label{table:speed}
$\textbf{Computational complexity.}$ 
Computational times ($\mu $s) for processing a single event with our EventNet and PointNet \cite{qi2017pointnet} are shown. 
To compute this statistic, we considered the case of an event rate of $1$ MEPS, which approximately corresponds to the highest event-rate scenario in the ETHTED+ dataset.
The spatial position of the synthetic event  was generated randomly and was temporarily uniform at the rate.
The statistic was  measured by a single core 3.2GHz Intel Core-i5.
The temporal window size of  $\tau=32$ ms is assumed.
To update the global feature $s_j$ using a newly received event $e_j$, mlp1/2 of PointNet needed to process a stream $\{\mathbf{e}_{j}|i=j-n(j)+1,...,j\}$ where $n(j)$ can be thousands or tens of thousands.
Similarly, $\max$ needs to process the $n(j)$ high-dimensional vector.
Conversely, input to EventNet was a single event $e_{j}$ because of the recursive formula of \Eq \ref{eq:recursive_2}.
Furthermore deep MLP is replaced with an LUT, which results in an extremely fast computation time of processing a single event in about 1 $\mu$s.
Thus, it can process (update global feature $s_j$) events of , at most, $1$ MEPS. 
The computation time for mlp3/mlp4 is less than  1 ms  meaning the application can query  the results at more than 1000 Hz.
We also report the computation time of naive mlp1/2 in parentheses for EventNet and observed that the processing of the MLP was accelerated about $45\times$ by the LUT.
}

\begin{center}
\begin{tabular}{c||c|c|c||c|c|c||c|c}
\hline 
\multicolumn{2}{c|}{} & \#input mlp1 & \#input $max$ & mlp1/2 & max pool(+t-code) & total & mlp3  & mlp4\tabularnewline
\hline 
\multicolumn{2}{c|}{PointNet} & $n(j)$ & $n(j)$ & $936.9\times10^{3}$ & $16.47\times10^{3}$ & $953.3\times10^{3}$ & $\textbf{0.58\ensuremath{\times\mathbf{10^{3}}}}$ & $0.59\times10^{3}$\tabularnewline
\hline 
\multicolumn{2}{c|}{EventNet } & $1$ & $2$ & $\textbf{0.65}(29.27)$  & $\textbf{0.36}$ & $\textbf{1.01}$ & $0.61\times10^{3}$ & $0.61\times10^{3}$\tabularnewline
\hline 
\end{tabular}
\end{center}
\end{table*}


%% file: fig/_batch.tex
\begin{figure}[!h]
\includegraphics[width=1\columnwidth]{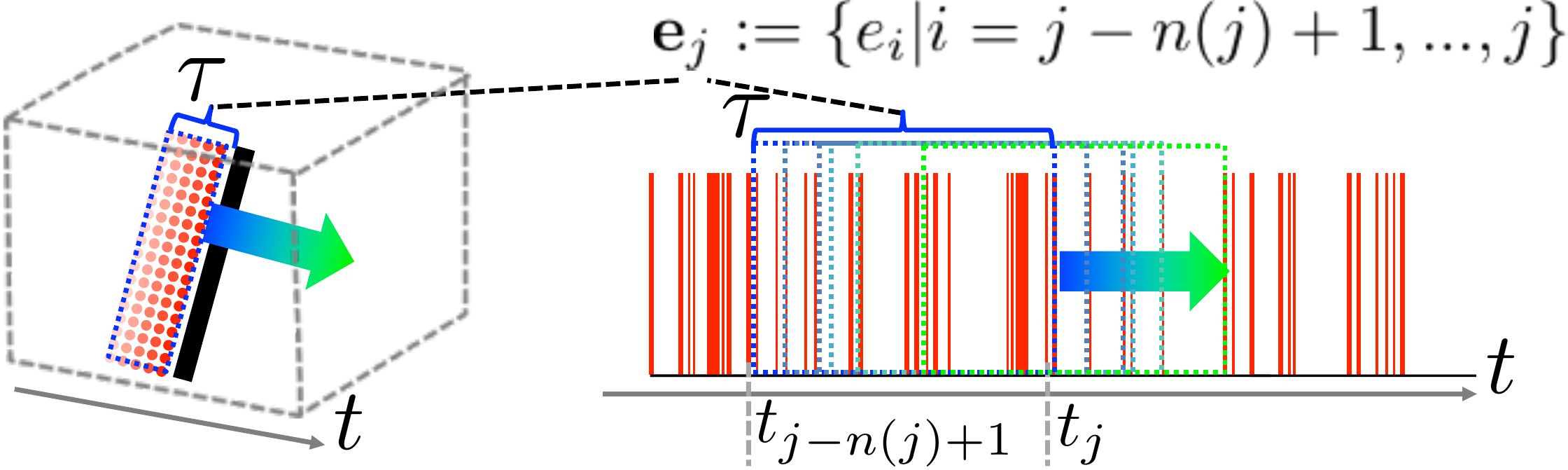}
\caption{
\label{fig:batch}
\textbf{Event data composition}
Left: An event stream from moving rod-like objects is illustrated.
Intensity changes caused by the movements of the objects generated an event stream.
Within a short period of time, different pixels of the sensor detect the intensity changes almost simultaneously.
Thus permutation between events happens even if the camera captures the same scene.
Right: We considered the event-stream $\mathbf{e}^t:=\{e_i|i=n(j)+1,...,j\}$ within a fixed temporal window $\tau$.
The number of events in the stream, $n(j)$, changed dynamically.
}
\end{figure}

%% file: fig/_quolitative.tex
\begin{figure*}[!h]
\includegraphics[width=2\columnwidth]{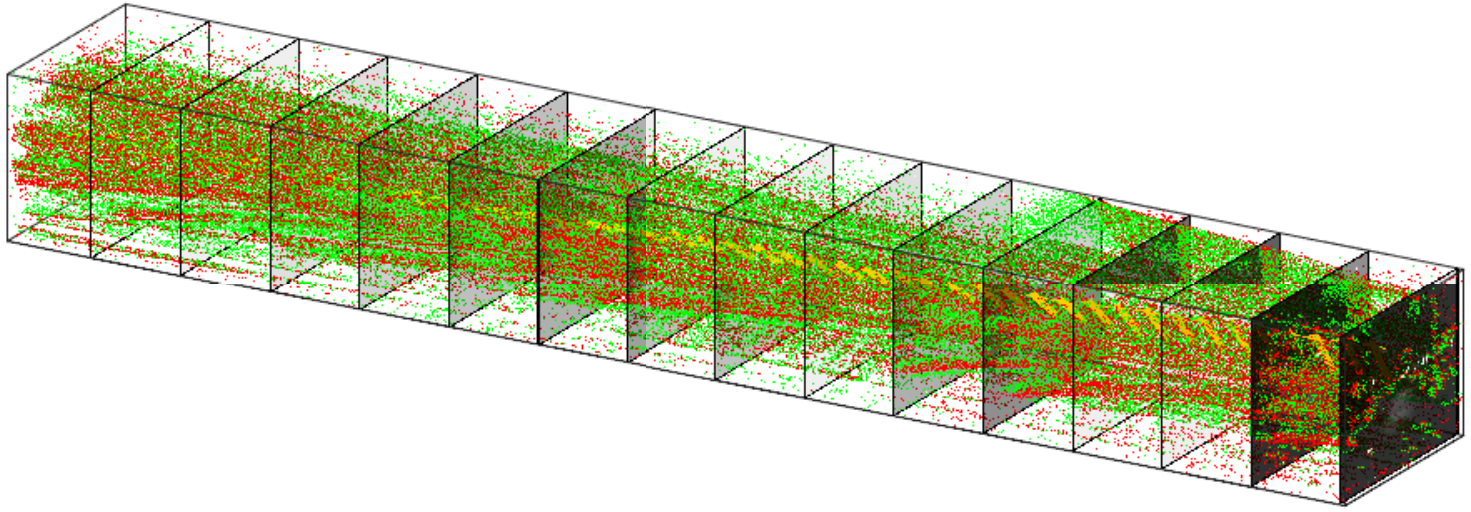}
\caption{
\label{fig:experiments_example}
\textbf{Snapshot from the MVSEC.}
Snapshot from the experiments using MVSEC for the ego-motion  (yaw rate) estimation task, where estimated  ego motions are shown as arrows (color encoded by the angle).
In this application, variable length event sequences are processed recursively in an event-wise manner updating the global feature at the variable event rate, and the ego motion is estimated at the rate of  1000 Hz using the system shown in \Fig \ref{fig:overview}.
Processing the input at this rate in real time is infeasible with the frame-based paradigm. 
}
\end{figure*}

%% file: section/related.tex
\section{Related Literature} 
\label{sec:bg}
\prg{Frame-Based DNNs for Modeling Event Data.}
There have been a few studies \cite{event_steer,zhu2018ev} attempting to model data from event cameras using DNNs.
The authors of \cite{event_steer} performed one of the pioneering works using CNNs to model event data.
In order to take advantage of existing 2D-CNN architecture, they converted raw spatiotemporal event data into an \textit{event frame} consisting of  2D histograms of positive and negative events.
The authors in  \cite{zhu2018ev} additionally concatenated \textit{time stamp images} to incorporate temporal information.
Most of the existing DNN-based approaches densify the sparse event signal to make use of the architecture of a frame-based paradigm such as with a CNN, which cannot make good use of the sparsity of the event stream.

\prg{Spiking Neural Networks}
Spiking neural networks (SNNs) \cite{HOTS,BMVC2016_94, bing2018ICRA,DBLP:journals/corr/TavanaeiM16b,mostafa2018supervised,Neftci2017,Lee2016,wang2015neuromorphic, Andreopoulos_2018_CVPR,10.3389/fnins.2017.00350,Davies2018} are third generation neural networks that are expected to process sparse asynchronous data efficiently.
The phased-LSTM \cite{NIPS2016_6310}  can also be interpreted as a kind of SNN. 
It is specifically designed to process asynchronous data such as event data in event-driven manner, 
and end-to-end supervised learning using a BP is possible similar to ours.
However, due to the architectural difference, its computational cost may be tens or hundreds of times more than our recursive LUT. 

\prg{Deep Learning on Unordered Sets}
PointNet \cite{qi2017pointnet} is a pioneering and successful method of dealing with unordered input sets, making use of a permutation invariant operation (such as $\max$) to deal with the unordered data in a concise and structured way.
PointNet and subsequent studies  \cite{qi2017pointnetplusplus,Zhou_2018_CVPR,Zhou_2018_CVPR} work remarkably well for many kind of tasks that require dealing with unordered point sets such as 3D point cloud data. 
However, since PointNet focuses on processing a set of points in a batch manner, its algorithm cannot process a sparse spatiotemporal event-stream recursively.

%% file: section/conclution.tex
\section{Conclusion}
\label{sec:conclusion}
We proposed EventNet, a trainable neural network designed for real-time processing of an asynchronous event stream in a recursive and event-wise manner.
We experimentally showed  usefulness for practical applications.
We will evaluate it in more challenging scenarios using the recently released event-camera simulator \cite{pmlr-v87-rebecq18a}. 
Our current architecture is the single layer of EventNet, but we will extend this work to a hierarchical structure such as the ones proposed in PointNet++ \cite{qi2017pointnetplusplus}. 
Another direction would be the its applications with LiDAR data \cite{Zhou_2018_CVPR,Zhou_2018_CVPR}, where we believe our model can be used to process point cloud data without waiting for the frame (360-degree rotation).